\documentclass[twocolumn]{rps-esrel2026}

\def\papername{\jobname}

\let\captionbox\relax
\usepackage{caption}
\usepackage{awesomebox}
\usepackage{enumitem}
\usepackage{natbib}
\usepackage{algorithm,algorithmic}
\usepackage{xcolor}
\usepackage{multirow}
\usepackage{tikz}
\usepackage{booktabs}
\usepackage{amsmath}
\usepackage{mathtools}

\begin{document}

\markboth{Ozan Kaya and Emir Cem Gezer}{Risk-Aware Obstacle Avoidance Algorithm for Real-Time Applications}

\twocolumn[

    \title{Risk-Aware Obstacle Avoidance Algorithm for Real-Time Applications}

    \author{Ozan Kaya}

    \address{Department of Marine Technology, Norwegian University of Science and Technology, Norway. \email{ozan.kaya@ntnu.no}}

    \author{Emir Cem Gezer}

    \address{Department of Marine Technology, Norwegian University of Science and Technology, Norway. \email{emir.cem.gezer@ntnu.no}}

    \author{Roger Skjetne}

    \address{Department of Marine Technology, Norwegian University of Science and Technology, Norway. \email{roger.skjetne@ntnu.no}}

    \author{Ingrid Bouwer Utne}

    \address{Department of Marine Technology, Norwegian University of Science and Technology, Norway. \email{ingrid.b.utne@ntnu.no}}

    \begin{abstract}
        Robust navigation in changing marine environments requires autonomous systems capable of perceiving, reasoning, and acting under uncertainty. This study introduces a hybrid risk-aware navigation architecture that integrates probabilistic modeling of obstacles along the vehicle path with smooth trajectory optimization for autonomous surface vessels. The system constructs probabilistic risk maps that capture both obstacle proximity and the behavior of dynamic objects. A risk-biased Rapidly Exploring Random Tree (RRT) planner leverages these maps to generate collision-free paths, which are subsequently refined using B-spline algorithms to ensure trajectory continuity. Three distinct RRT* rewiring modes are implemented based on the cost function: minimizing the path length, minimizing risk, and optimizing a combination of the path length and total risk. The framework is evaluated in experimental scenarios containing both static and dynamic obstacles. The results demonstrate the system's ability to navigate safely, maintain smooth trajectories, and dynamically adapt to changing environmental risks. Compared with conventional LIDAR or vision-only navigation approaches, the proposed method shows improvements in operational safety and autonomy, establishing it as a promising solution for risk-aware autonomous vehicle missions in uncertain and dynamic environments.
    \end{abstract}


]

\section{Introduction}

Autonomous Surface Vessels (ASVs) have been gaining a strong attention in the digital transformation of the marine sector, offering safer, cleaner, and more efficient operations \citep{RenanPHDthesis}. 
Recent advances in relevant technologies have significantly improved their capability to perform missions such as monitoring, inspection, and transportation. However, robust and safe navigation in real world remains a major research challenge.
The primary cause is because the consequences of hazardous in the marine domain can be very high.
To achieve higher levels of autonomy ASVs must not only detect and track obstacles but also reason about environmental uncertainty and act safely while considering the risk picture \citep{utne2020towards}.

\emph{Risk} in marine systems and operations encompasses a wide variety of design and operational hazards. For ASV relevant navigation hazards and hazardous events are collisions, groundings, and drift hazardouss \citep{haugen2022maritime}. Traditional \emph{risk assessment} frameworks, such as those discussed by \citet{rausand2020risk}, have typically focused on the design phase and assumed relatively static operating conditions. However, recent developments in dynamic and online Risk Assessment (RA) methodologies enable a more continuous updating of the risk spectrum \citep{KHAN20169, utne2017risk, Bremnes10704971}. Such approaches can be integrated directly into autonomous control loops which allows ASVs to adapt their mission dynamically in response to environmental and operational changes; see e.g. \citep{utne2020towards, kristensen2025evaluating}.

Probability theory is a key element of such dynamic risk assessment frameworks. One feasible method is Bayesian Belief Networks (BBNs) and related graphical models that provide a structured means of capturing dependencies between uncertain operational variables such as vessel velocity, obstacle proximity, control system failures, and environmental disturbances, e.g., \citep{maidana2023risk, johansen2022supervisory}.
By frequent updating of the conditional probabilities based on situational awareness, these models can estimate the likelihood or risk in operation.

Several studies have integrated accident probability estimations into path planning algorithms.
\citet{lefebvre2016integration} and \citet{chen2022risk} proposed \emph{risk-weighted} A* and D* methods for underwater and surface vehicle navigation, showing improved collision avoidance under environmental uncertainties.
\citet{zhen2023improved} extended this concept to ship navigation using a modified A* algorithm that considered both environmental hazards and vessel dynamics.
Similarly, \citet{FISKIN2021108502} employed fuzzy logic and genetic optimization to enhance risk-aware collision avoidance performance.
However, risk is typically included as a fixed cost or heuristic, which reduces the transparency of its interpretation and adaption during planning and drives the need for more open formulations that directly address risk \citet{Utne2025onRiskMetrics}.
Consequently, there is a need for integrated frameworks that combine probabilistic reasoning, adaptive risk modeling, and continuous trajectory optimization for real-world applications.

This study accounts for identified hazardous events by estimating their probabilities of occurrence using a Bayesian Belief Network (BBN), resulting in a multilayered risk-aware navigation architecture that integrates probabilistic risk modeling with smooth trajectory optimization.
The system constructs probabilistic risk maps by fusing environmental observations (e.g., radar, LIDAR, and tracking data) with predicted behaviors of dynamic obstacles and exploring the navigation space.
The resulting hazardous event probability map captures both static and dynamic variations and is used for navigation decisions  \citet{kristensen2025evaluating}.
The path planning is achieved using a Risk-Aware Rapidly Exploring Random Tree (RA-RRT*) algorithm that incorporates the probabilistic risk field directly into its sampling and cost functions, considering the advantages of RRT* as investigated by \citet{ozan10708298}.

To this end, we have developed a RA-RRT* path planning algorithm that incorporates hazardous probability estimations from a BBN and uses a weighting value to balance risk probability and length along the path.
This enables flexible trajectory generation that can be tuned according to mission context, risk tolerance, or environmental conditions \citep{utne2020towards, vagale2021bpath}. We then generate smooth trajectories from initially sampled waypoints using B-splines which guarantees curvature continuity and eliminates sharp turning angles.



This work contributes to a risk-aware obstacle avoidance framework that bridges probabilistic risk modeling with adaptive trajectory planning for ASVs. By integrating a BBN model, RRT*, and smooth trajectory optimization, the system achieves safer operation in dynamic marine environments.

This paper is structured as follows: Section \ref{sec:preliminaries} lays down the foundation and talks about the preliminaries. Section \ref{section:methodology} offers a methodology, including an overview of the proposed framework, the BBN model, and RA-RRT* path planning for the trajectory. Furthermore, this section presents evaluation scenarios according to different cost function and rewiring methods. Section \ref{sec:result} presents the results and discussions, followed by a conclusion summarizing the key findings and outlining avenues for future research.

\section{Preliminaries}
\label{sec:preliminaries}

\subsection{Risk definition}

In this work, risk is represented as a collection of triplets describing hazardous events, their probabilities, and consequences \citep{kaplan1981quantitative}.
The notation follows as presented by \citet{lefebvre2016integration}.
Let the set of hazardous events be denoted as
\begin{equation}
    R = \left\{  E_i,\ \Pr\{E_i\},\ C_i \right\}_{i=1}^{n_R},
    \label{eq:Risk}
\end{equation}
where $E_i$ represents hazardous event $i$, $\Pr\{E_i\}$ is the probability of that event, and $C_i$ denotes the consequences. When the set includes all relevant hazardous events identified through a systematic risk management process, it is considered to provide a complete representation of risk.

\subsection{Bayesian Belief Networks}

BBNs provide a graphical probabilistic representation of causal dependencies affecting collision or grounding. Each node represents an influencing factor. This model enables reasoning under uncertainty by continuously updating belief states when new evidence becomes available.

\subsection{Problem Definition}

Let the path planning problem be defined on a directed graph $\mathcal{G = (V, A)}$, where $\mathcal{V}$ denotes the set of nodes and $\mathcal{A \subseteq V \times V}$ denotes the set of directed edges. A path $p$ is defined as an ordered sequence of edges connecting a start node $x_s \in V$ to a goal node $x_g \in V$. Let $\mathcal{P}(x_s, x_g)$ denote the set of all feasible paths from $x_s$ to $x_g$. For a given path $p \in \mathcal{P}(x_s, x_g)$, $\mathcal{A}(p) \subseteq \mathcal{A}$ denote the set of edges traversed by $p$. Finally, let $\Pr\{E_i \mid a\}$ denote the probability of hazardous event $E_i$ occurring while traversing edge $a(u,v) \in \mathcal{A}$.
Therefore we write,
\begin{align}
    h(a) & \coloneq \sum_{i=1}^{n_R} \Pr\{E_i \mid a\},     \\
    c(a) & \coloneq \sum_{i=1}^{n_R} C_i \Pr\{E_i \mid a\},
\end{align}
where $h(a)$ is hazard probability and $c(a)$ is cost up until the edge $a$.
The path-wise hazardous event probability and consequence functions $C, H: \mathcal{P}(x_s, x_g) \to \mathbb{R}_{\ge 0}$ is therefore can be defined as
\begin{align}
    C(p) \coloneq \sum_{a \in \mathcal{A}(p)} c(a), \, H(p) \coloneq \sum_{a \in \mathcal{A}(p)} h(a).
\end{align}
For any edge $a = (u, v) \in A$, let $d(a)$ denote the Euclidean distance between the states associated with nodes $u$ and $v$, that is
\begin{align}
    d(a) = \|\mathbf{x}_u - \mathbf{x}_v\|_2.
\end{align}
Then, the total path distance $D(p)$ is the sum of these edge lengths,
\begin{align}
    D(p) \coloneq \sum_{a \in A(p)} d(a).
\end{align}
The baseline objective function $J(p)$ is then expressed as a combination of this geometric distance and a risk bias term $B(p)$, according to
\begin{align}
    J(p) = D(p) + B(p).
\end{align}
The sub-optimal path $p^*$ is determined by minimizing a composite cost function that balances geometric efficiency and path-wise hazardous event probability, that is,
\begin{align}
    p^* = \arg \min_{p \in \mathcal{P}(x_s, x_g)} J(p),
\end{align}
where different instantiations of the bias term $B(p)$ yield distinct risk-aware planning behaviors.

\section{Methodology}
\label{section:methodology}

\subsection{Overview of the Proposed Framework}
\label{section:methodology:overview}

\begin{figure}[t]
    \centering
    \includegraphics[width=0.95\linewidth, page=1]{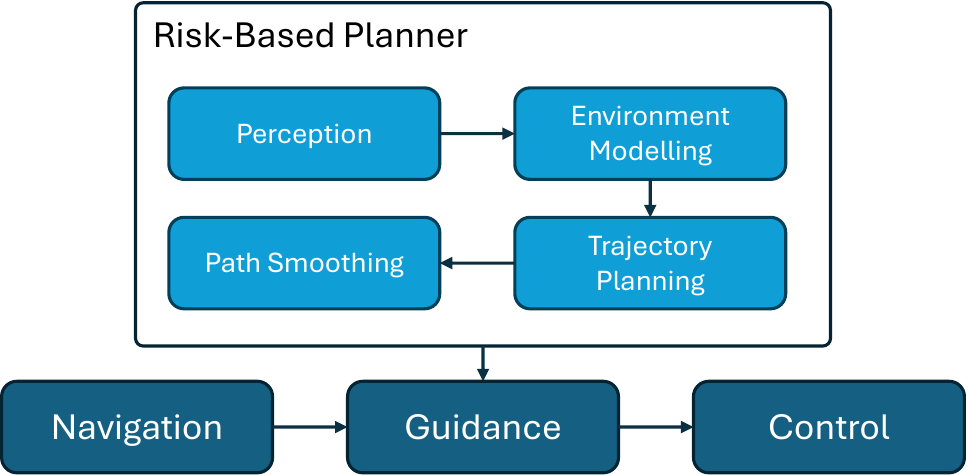}
    \caption{System architecture.}
    \label{fig:framework_overview}
\end{figure}

The proposed framework integrates probabilistic risk modeling with trajectory generation.
The architecture consists of four main modules: perception, probabilistic environmental and risk modeling, RA-RRT* path planning, and path smoothing.
A overview of the proposed framework is illustrated in Fig. \ref{fig:framework_overview}.

The framework processes obstacle information and other relevant measurements to set evidences of the BBN to compute the risk probabilities.
The computed probabilities are then provided to the RA-RRT* planner, where probabilities directly influence node costs and rewiring decisions.
This enables the planner to avoid high-risk regions.
The resulting discrete waypoint sequence is then passed to the trajectory smoothing module, where a B-spline algorithm is used for constructing a parametrically continuous curve.

Running RA-RRT* produces nodes with hazardous event probabilities and total distance associated with them.
Using the event probabilities, we create a risk map that shows the areas of high hazardous probabilities by interpolating node probability values.
One run of RA-RRT* creates a lookup table with nodes with precomputed hazardous event probabilities and total distance from initial position to goal position.

\subsection{BBN model}
\label{sec:methodology:bbn_model}

An important aspect of risk modeling is the identification of Risk-Influencing Factors (RIFs), which are system properties, environmental conditions, or operational states that affect either the likelihood or consequences of hazardous events \citep{rausand2020risk}.
In the context of autonomous marine vehicles, RIFs are closely related to system dynamics, obstacle encounters, environmental conditions, and system performance.
Among the potential hazardous events, collision and grounding are relevant for path planning and obstacle avoidance.

The BBN used in this work is taken from the online navigation risk assessment model proposed by \citet{maidana2023risk}, which estimates the probability of navigation accidents during ASV operations.
For simplicity we have only provided evidence to select number of nodes.
These nodes are \emph{distance to Shore}, \emph{depth}, Distance to Closest Point of Approach \emph{DCPA}, \emph{target ship on collision course}, and \emph{vessel on grounding course}.

We consider grounding and collision events probabilities, such that $n_R = 2$.
The resulting risk probabilities are used as inputs to the RA-RRT* planner, where risk is treated as a term influencing rewiring and path generation.

The network considers factors related to encounter properties, vessel maneuverability, equipment reliability, and environmental conditions.
These factors are discretized into qualitative or quantitative states and connected to the accident node through a set of Conditional Probability Tables (CPTs) that quantify their respective influence.
We used a BBN model implemented using the PySMILE Python interface to the GeNIe reasoning engine \cite{druzdzel1999smile}, which performs iterative updates within the proposed RA-RRT* planner.

\subsection{Risk-Aware RRT* Path Planning}
\label{sec:methodology:planner}

Conventional RRT and its optimized variant RRT* are among the most widely used algorithms for sampling-based motion planning in high-dimensional spaces \citep{ozan10708298}.
These algorithms efficiently explore configuration spaces by incrementally constructing a tree rooted at the start position and connecting randomly sampled states to nearby existing nodes through a steering function.
While RRT* guarantees asymptotic optimality in terms of path length, it does not inherently account for operational risk, making it unsuitable for risk-critical marine applications where probabilistic risk from dynamic obstacles, waves, or navigation errors are present.

To address these limitations, the RA-RRT* algorithm is proposed in this study, which integrates probabilistic risk modeling directly into the node sampling, cost evaluation, and rewiring mechanisms.
The rewiring process allows the planner to gradually converge toward an optimal solution rather than settling for the first feasible path.
The key modification lies in the cost function, formulated as given below,
\begin{gather}
    J=\alpha   D(p) + (1-\alpha) \beta B(p),
    \label{eq:cost}
\end{gather}
where
$\beta$ is normalization coefficient, and $\alpha$ is weighting with balancing between geometric optimality and risk.
When $\alpha=1$, the planner behaves like a classical RRT*, focusing purely on shortest-path optimization. Conversely, when $\alpha=0$, the algorithm only uses risk.



\begin{algorithm}[!t]
    \caption{Risk-Aware RRT* Path Planning}
    \begin{algorithmic}[1]
        \label{RA-RRTalgorithmcode}
        \renewcommand{\algorithmicrequire}{\textbf{Input:}}
        \renewcommand{\algorithmicensure}{\textbf{Output:}}
        \REQUIRE Start node $x_{start}$, goal region $X_{goal}$, risk map $R(x)$, configuration space $X_{free}$, weighting coefficient $\alpha$, no. of iterations $N$
        \ENSURE  Risk-aware collision-free trajectory $P^*$
        \\ \STATE $T.\text{Initialize}(x_{start})$
        \FOR{$i=1$ to $N$}
        \STATE $x_{rand} \gets \text{Sample}(X_{free}, {B}(x))$
        \STATE $x_{near} \gets \text{Nearest Neighbor}(T, x_{rand})$
        \STATE $x_{new} \gets \text{Steer}(x_{near}, x_{rand})$
        \IF{\textbf{not} \text{Collision Free}$(x_{near}, x_{new}, {B}(x))$}
        \STATE \textbf{continue}
        \ENDIF
        \STATE $X_{near} \gets T.\text{Neighboring Nodes}(x_{new})$
        \STATE $x_{parent} \gets \text{First Element in } X_{near}$
        \FOR{$\textbf{each} \ x_i \in X_{near}$}
        \STATE ${D} \gets \text{Path Length}(x_i, x_{new})$
        \STATE ${B} \gets \text{Integrate Risk}({B}(x), x_i, x_{new})$
        \STATE $J_{new} \gets \alpha \ {D} + (1 - \ \alpha) \ \beta \ {B}$
        \IF{$J_{new} < \text{Cost}(x_{parent}, x_{new})$}
        \STATE $x_{parent} \gets x_i$
        \ENDIF
        \ENDFOR
        \STATE $T.\text{Add Node}(x_{new})$
        \STATE $T.\text{Add Edge}(x_{parent}, x_{new})$
        \FOR{$\textbf{each} \ x_i \in x_{near}$}
        \STATE $ {D} \gets \text{distance}(x_{new}, x_i)$
        \STATE $ {D} \gets \text{risk}({D}(x), x_{new}, x_i)$
        \STATE $J_{rewire} \gets \alpha \ {D} + (1 - \ \alpha) \ \beta \ {B}$
        \IF{$J_{rewire} < \text{Cost}(x_{parent}, x_i)$}
        \STATE $T.\text{Rewire}(x_{new}, x_i)$
        \ENDIF
        \ENDFOR
        \ENDFOR
        \STATE $P^* \gets T.\text{Extract Best Path}(X_{goal})$
        \RETURN $P^*$
    \end{algorithmic}
\end{algorithm}
Dynamic obstacles and time-varying risks are handled through risk-aware rewiring, where hazardous event probability is estimated by a BBN that is integrated into the rewiring cost function. At each simulation step, the algorithm updates the risk field and invalidates edges that intersect newly detected high-risk regions.

\begin{figure*}[htbp]
    \centering
    \includegraphics[width=\linewidth]{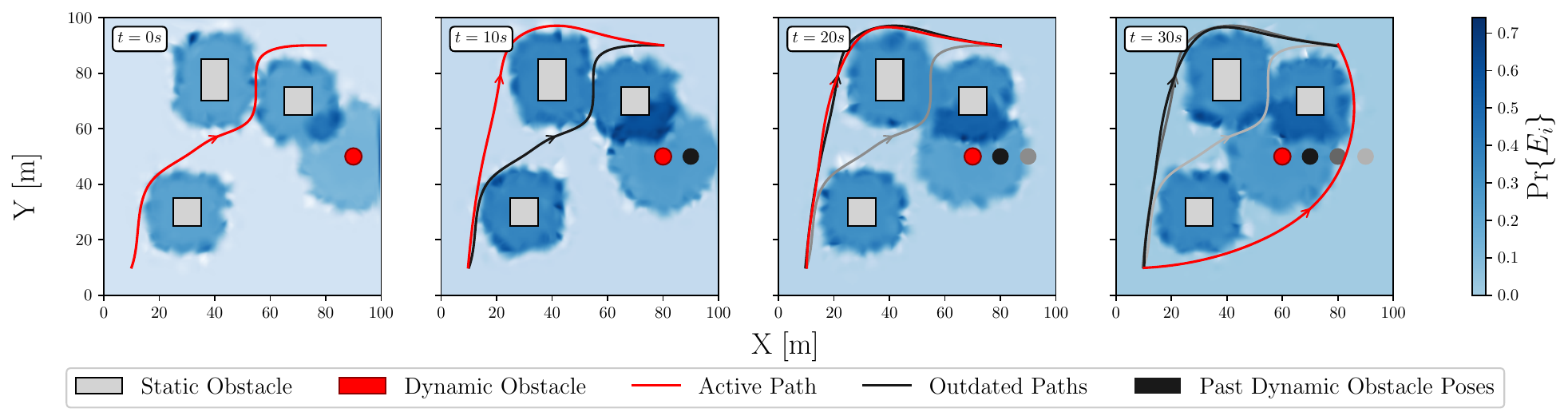}
    \caption{
        Path planning result as the dynamic obstacle moves towards the center of the search space. The path is computed with $\alpha=0.2$. Path is replanned every $10$ seconds.
    }
    \label{fig:dynamic_obstacle_replanning_results}
\end{figure*}

\section{Results and Discussion}
\label{sec:result}

This section presents the performance evaluation of the proposed RA-RRT* framework in a simulated marine environment containing both static and dynamic obstacles.
The experiments were conducted using three planning configurations: (1) Shortest-path mode ($\alpha = 1$), prioritizing minimum Euclidean distance; (2) risk prioritizing mode ($\alpha = 0$), prioritizing minimum aggregated risk probability; and (3) combined risk-distance mode ($0 < \alpha < 1$).
In Fig. \ref{fig:dynamic_obstacle_replanning_results} shows the path planning result as the dynamic obstacle moves towards the center of the search space. The path is computed with $\alpha=0.2$.
As the dynamic obstacle moves, BBN inputs are updated accordingly, and the risk map is revised to reflect the changing hazardous event probabilities.

\begin{figure}[htb]
    \centering
    \includegraphics[width=\linewidth]{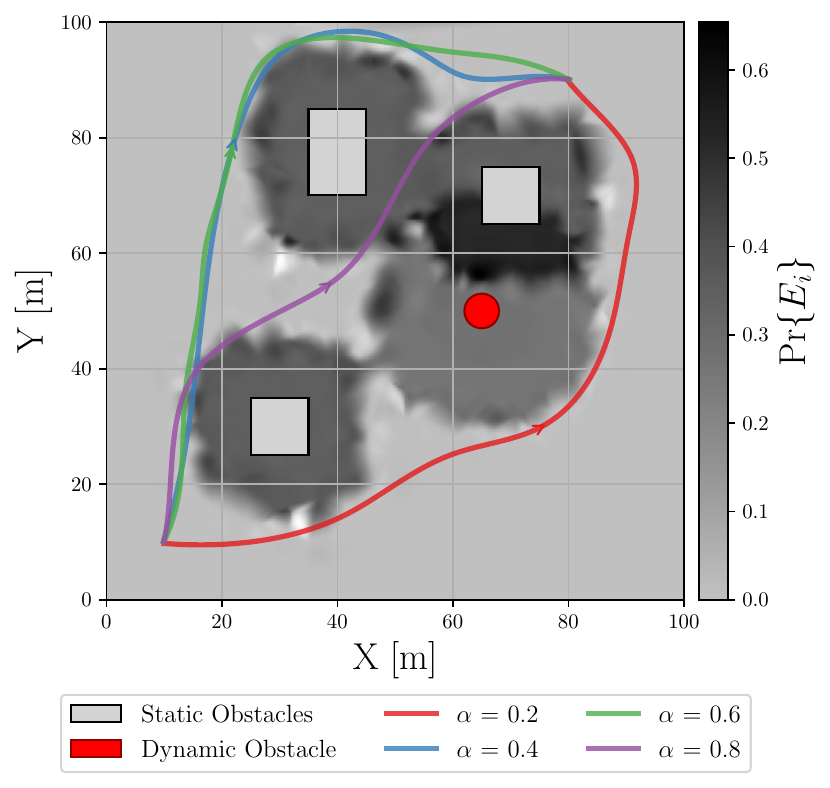}
    \caption{The replanning path results in different value of weighting-coefficient ($\alpha$).}
    \label{fig:alphas}
\end{figure}

The shortest-path mode consistently generated the smallest path lengths, as expected, but also produced the highest hazardous event probability values, particularly in proximity to both dynamic and static obstacles. To illustrate the rewiring results, the paths generated in different value of weighting-coefficient ($\alpha$). Rewiring path results related to $\alpha$ are displayed in Fig. \ref{fig:alphas}.

The risk prioritizing mode yielded significantly lower risk probabilities, often by an order of magnitude compared with the shortest-path approach. However, this came with increased path length due to conservative detours around the points which indicates high collision and grounding probability.

Hybrid $\alpha$-weighted planning demonstrated balanced behavior, achieving substantial hazardous event probability reduction relative to $\alpha = 0$ while maintaining moderate path lengths compared to the $\alpha = 1$ strategy.
Table \ref{tab:alpha_results} summarizes the results from $N=50$ independent trials for each $\alpha$. The risk here corresponds to the cumulative BBN-derived hazardous event probability related to collision and grounding along the path, while the path length is measured after B-spline smoothing.

\begin{table*}[ht!]
    \centering
    \caption{Statistics of $D(p^*)$ and $\Pr\{E_i\}$ for different values of $\alpha$.}
    \label{tab:alpha_results}
    \begin{tabular}{c c c c c c}
        \toprule
        $\alpha$ & $D(p^*)$ Mean          & $D(p^*)$ Std           & $\Pr\{E_i\}$ Mean      & $\Pr\{E_i\}$ Std       & $\Pr\{E_i\}$ Max       \\
        \midrule
        0.1      & $1.416\mathrm{e}{+02}$ & $1.634\mathrm{e}{+00}$ & $0.000\mathrm{e}{+00}$ & $0.000\mathrm{e}{+00}$ & $0.000\mathrm{e}{+00}$ \\
        0.3      & $1.181\mathrm{e}{+02}$ & $5.201\mathrm{e}{+00}$ & $4.102\mathrm{e}{-02}$ & $9.666\mathrm{e}{-02}$ & $4.985\mathrm{e}{-01}$ \\
        0.5      & $1.150\mathrm{e}{+02}$ & $2.088\mathrm{e}{+00}$ & $5.390\mathrm{e}{-02}$ & $1.026\mathrm{e}{-01}$ & $4.985\mathrm{e}{-01}$ \\
        0.7      & $1.114\mathrm{e}{+02}$ & $1.453\mathrm{e}{+00}$ & $9.414\mathrm{e}{-02}$ & $1.265\mathrm{e}{-01}$ & $5.115\mathrm{e}{-01}$ \\
        0.9      & $1.090\mathrm{e}{+02}$ & $1.038\mathrm{e}{+00}$ & $1.536\mathrm{e}{-01}$ & $1.561\mathrm{e}{-01}$ & $5.115\mathrm{e}{-01}$ \\
        \bottomrule
    \end{tabular}
\end{table*}

\begin{figure}[th]
    \centering
    \includegraphics[width=\linewidth]{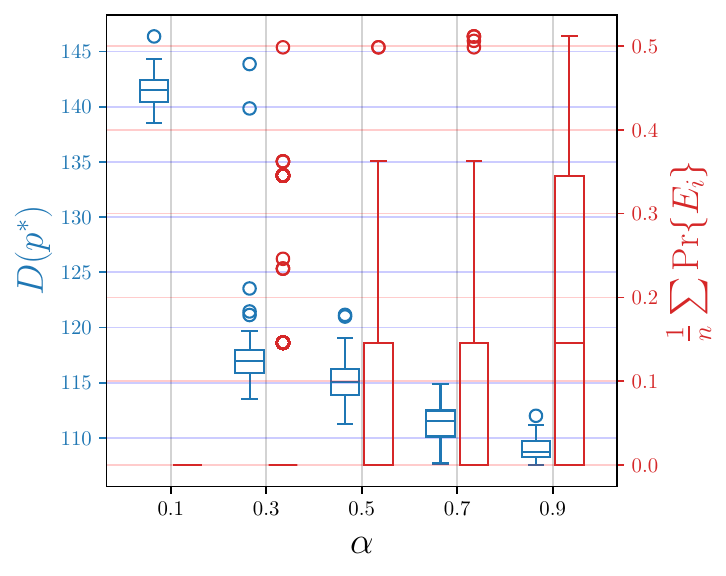}
    \caption{Box plots of path length $D(p^*)$ and hazardous event probability $\frac{1}{n} \sum \Pr\{E_i\}$ for different values of $\alpha$.}
    \label{fig:placeholder}
\end{figure}

Due to the probabilistic sampling nature of RA-RRT*, multiple feasible paths can be generated even under identical initial conditions. This characteristic is advantageous in marine scenarios where obstacle motion and uncertainty evolve over time. Figure \ref{fig:placeholder} shows box plot related to path length $D(p*)$ and hazardous event probability $Pr\{E_i\}$ for chosen $\alpha$ values.

In shortest-path mode, variability between trials remained low, producing highly consistent trajectories concentrated around a single corridor. In risk prioritizing mode, higher variability was observed, as the planner actively avoids dynamically changing high hazardous event probability areas computed by the BBN.

The qualitative results show that the RA-RRT* strongly adapts to changes in dynamic obstacle behavior. This leads to: (1) Suppression of node expansion in high-risk regions; (2) Rewiring of existing tree branches to avoid rising-risk areas; and (3) Shift of sampling density toward safer zones, improving both path feasibility and consistency.

Compared with a conventional RRT*, which reacts only to geometric constraints, such as distance, RA-RRT* achieves more proactive avoidance by considering collision and grounding through the integrated BBN.

\section{Conclusion}
\label{sec:conclusion}

This paper presented a hybrid risk-aware navigation framework for autonomous surface vessels that integrates Bayesian risk modeling with an RRT* path planner.
By combining probabilistic risk maps, generated through a BBN, with a sampling-based planner, the proposed method enables autonomous vessels to perceive, reason, and act effectively under uncertainty.
The three planning modes introduced in this work, shortest-path, risk-minimizing, and $\alpha$-weighted planning, allow mission designers to balance operational efficiency and risk according to situational requirements.

Results demonstrate that the RA-RRT* algorithm successfully avoids both static and dynamic obstacles while proactively steering away from regions with considering probabilistic risk.
Compared with traditional path planning approaches (shortest path), the risk-aware planner achieved  reduced cumulative hazardous event probability, particularly in environments with moving objects.
The $\alpha$-weighted planner enabled flexible trade-offs between path length and risk, producing trajectories that adapt smoothly to dynamic and static obstacle conditions. The integration of B-spline smoothing ensured that the final trajectories remained smooth and feasible.

Overall, the proposed framework provides a useful foundation for further development of safe autonomous navigation in uncertain marine environments.
The results show that combining probabilistic risk inference with sub-optimal sampling-based planning enhances both robustness and situational awareness, outperforming geometry-only navigation strategies. This sub-optimality, however, also affects risk, since the selected trajectory is not guaranteed to minimize risk globally.
Despite the promising results, the proposed risk metric is here using a simplified formulation and relies on a BBN that incorporates only a limited set of RIFs that do not fully capture the complexity of real-world operational uncertainty.
Although operational constraints can significantly affect risk acceptance, their impact is not explicitly considered in this work.
Incorporating constraint-driven risk acceptance and adaptive risk thresholds remains an important direction for future research.

Future work will focus on extending this framework to real-time onboard operation, including more RIFs, validating the approach on physical autonomous surface vessel platforms, and incorporating additional sensing modalities to further improve environmental understanding and prediction accuracy.

\begin{acknowledgement}
    Kaya and Utne´s contributions are funded by the European Union and the ERC grant (BREACH, grant id. 101142277, DOI 10.3030/101142277). Views and opinions expressed are however those of the author(s) only and do not necessarily reflect those of the European Union or the European Research Council Executive Agency.
    Neither the European Union nor the granting authority can be held responsible for them.
    Gezer and Skjetne´s contributions are funded by the SFI AutoShip (RCN grant id. 309230), and partly by the Norwegian Maritime AI Centre (RCN grant id. 359242).
    \vspace{-1cm}
    \begin{figure}
        \includegraphics[width=0.8\linewidth]{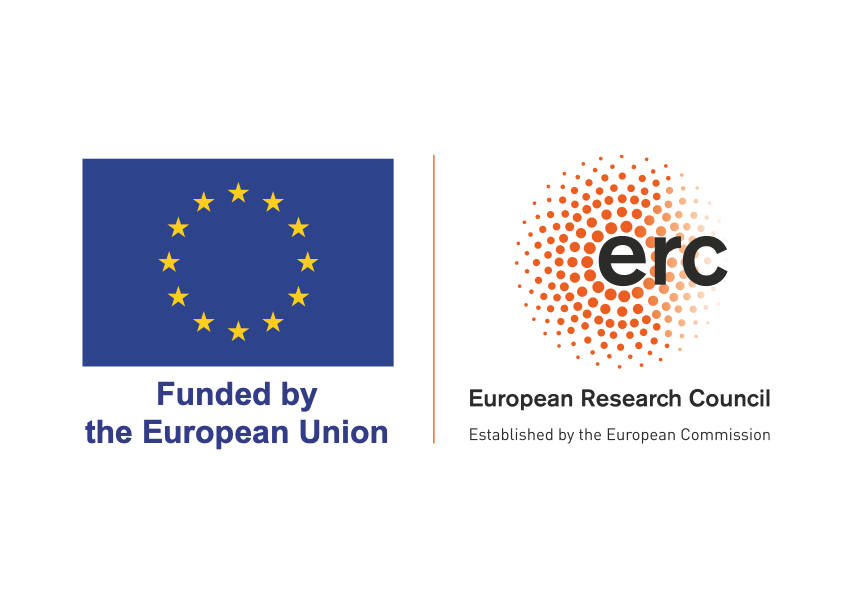}
        \label{fig:fundation}
    \end{figure}
\end{acknowledgement}
\vspace{-1cm}
\bibliographystyle{chicago}
\bibliography{References}

@article{johansen2022supervisory,
  title   = {Supervisory risk control of autonomous surface ships},
  author  = {Johansen, Thomas and Utne, Ingrid Bouwer},
  journal = {Ocean Engineering},
  volume  = {251},
  pages   = {111045},
  year    = {2022}
}

@book{haugen2022maritime,
  title     = {Maritime transportation: safety management and risk analysis},
  author    = {Haugen, Stein and Kristiansen, Svein},
  year      = {2022},
  publisher = {Routledge}
}

@inproceedings{utne2017risk,
  title        = {Risk management of autonomous marine systems and operations},
  author       = {Utne, Ingrid Bouwer and S{\o}rensen, Asgeir J and Schj{\o}lberg, Ingrid},
  booktitle    = {International conference on offshore mechanics and arctic engineering},
  volume       = {57663},
  pages        = {V03BT02A020},
  year         = {2017},
  organization = {American Society of Mechanical Engineers}
}

@article{utne2020towards,
  title   = {Towards supervisory risk control of autonomous ships},
  author  = {Utne, Ingrid Bouwer and Rokseth, B{\o}rge and S{\o}rensen, Asgeir J and Vinnem, Jan Erik},
  journal = {Reliability Engineering \& System Safety},
  volume  = {196},
  pages   = {106757},
  year    = {2020}
}

@article{lefebvre2016integration,
  title   = {Integration of risk in hierarchical path planning of underwater vehicles},
  author  = {Lefebvre, Nicolas and Schj{\o}lberg, Ingrid and Utne, Ingrid B},
  journal = {IFAC-PapersOnLine},
  volume  = {49},
  number  = {23},
  pages   = {226--231},
  year    = {2016}
}

@article{chen2022risk,
  title   = {Risk-based path planning for autonomous underwater vehicles in an oil spill environment},
  author  = {Chen, Xi and Bose, Neil and Brito, Mario and Khan, Faisal and Millar, Gina and Bulger, Craig and Zou, Ting},
  journal = {Ocean Engineering},
  volume  = {266},
  pages   = {113077},
  year    = {2022}
}

@article{zhen2023improved,
  title   = {An improved A-star ship path-planning algorithm considering current, water depth, and traffic separation rules},
  author  = {Zhen, Rong and Gu, Qiyong and Shi, Ziqiang and Suo, Yongfeng},
  journal = {Journal of Marine Science and Engineering},
  volume  = {11},
  number  = {7},
  pages   = {1439},
  year    = {2023}
}

@article{maidana2023risk,
  title   = {Risk-based path planning for preventing collisions and groundings of maritime autonomous surface ships},
  author  = {Maidana, Renan Guedes and Kristensen, Susanna Dybwad and Utne, Ingrid Bouwer and S{\o}rensen, Asgeir Johan},
  journal = {Ocean Engineering},
  volume  = {290},
  pages   = {116417},
  year    = {2023}
}

@article{KHAN20169,
  title   = {Dynamic risk management: a contemporary approach to process safety management},
  journal = {Current Opinion in Chemical Engineering},
  volume  = {14},
  pages   = {9-17},
  year    = {2016},
  issn    = {2211-3398},
  doi     = {https://doi.org/10.1016/j.coche.2016.07.006},
  url     = {https://www.sciencedirect.com/science/article/pii/S2211339816300399},
  author  = {Faisal Khan and Seyed Javad Hashemi and Nicola Paltrinieri and Paul Amyotte and Valerio Cozzani and Genserik Reniers}
}

@article{Bremnes10704971,
  author   = {Bremnes, Jens Einar and Utne, Ingrid Bouwer and Krogstad, Thomas Røbekk and Sørensen, Asgeir Johan},
  journal  = {IEEE Journal of Oceanic Engineering},
  title    = {Holistic Risk Modeling and Path Planning for Marine Robotics},
  year     = {2025},
  volume   = {50},
  number   = {1},
  pages    = {252-275},
  doi      = {10.1109/JOE.2024.3432935}
}

@article{FISKIN2021108502,
  title    = {Fuzzy domain and meta-heuristic algorithm-based collision avoidance control for ships: Experimental validation in virtual and real environment},
  journal  = {Ocean Engineering},
  volume   = {220},
  pages    = {108502},
  year     = {2021},
  issn     = {0029-8018},
  doi      = {https://doi.org/10.1016/j.oceaneng.2020.108502},
  url      = {https://www.sciencedirect.com/science/article/pii/S0029801820314098},
  author   = {Remzi Fiskin and Oguz Atik and Hakki Kisi and Efendi Nasibov and Tor Arne Johansen}
}

@article{vagale2021bpath,
  title   = {Path planning and collision avoidance for autonomous surface vehicles II: a comparative study of algorithms},
  author  = {Vagale, Anete and Bye, Robin T and Oucheikh, Rachid and Osen, Ottar L and Fossen, Thor I},
  journal = {Journal of Marine Science and Technology},
  volume  = {26},
  number  = {4},
  pages   = {1307--1323},
  year    = {2021}
}

@inproceedings{ozan10708298,
  author    = {Kaya, Ozan and Tingelstad, Lars},
  booktitle = {2024 10th International Conference on Control, Decision and Information Technologies (CoDIT)},
  title     = {Comparison of RRT, APF, and PSO-Based RRT-APF (PS-RRT-APF) for Collision-Free Trajectory Planning in Robotic Welding},
  year      = {2024},
  volume    = {},
  number    = {},
  pages     = {2639-2644},
  doi       = {10.1109/CoDIT62066.2024.10708298}
}

@inproceedings{druzdzel1999smile,
  title     = {SMILE: Structural Modeling, Inference, and Learning Engine and GeNIe: a development environment for graphical decision-theoretic models},
  author    = {Druzdzel, Marek J},
  booktitle = {Aaai/Iaai},
  pages     = {902--903},
  year      = {1999}
}

@article{kaplan1981quantitative,
  title   = {On the quantitative definition of risk},
  author  = {Kaplan, Stanley and Garrick, B John},
  journal = {Risk analysis},
  volume  = {1},
  number  = {1},
  pages   = {11--27},
  year    = {1981}
}

@article{kristensen2025evaluating,
  title   = {Evaluating the effect of risk metrics for supporting operational decision-making by autonomous surface vehicles},
  author  = {Kristensen, Susanna D and Maidana, Renan G and Utne, Ingrid B and Bremnes, Jens E},
  journal = {Ocean Engineering},
  volume  = {338},
  pages   = {121937},
  year    = {2025}
}

@book{rausand2020risk,
  title     = {Risk assessment: theory, methods, and applications (2020 ed.)},
  author    = {Rausand, Marvin and Haugen, Stein},
  year      = {2020},
  publisher = {John Wiley \& Sons}
}

@inproceedings{Utne2025onRiskMetrics,
author = {Utne, Ingrid},
year = {2025},
month = {01},
pages = {2423-2430},
title = {A Systematic Review of Risk Metrics for AI and Autonomous Systems},
booktitle    = {35th European Safety and Reliability Conference (ESREL 2025)},
doi = {10.3850/978-981-94-3281-3_ESREL-SRA-E2025-P7137-cd}
}

@phdthesis{RenanPHDthesis,
    author = {Maidana, Renan Guedes},
    title = {Risk Assessment for DecisionMaking in Autonomous Marine
and Maritime Systems},
    school = {Norwegian University of Science and Technology},
    year = {2024}
}

\end{document}